\pdfoutput=1

\documentclass[11pt]{article}

\usepackage[final]{acl}

\usepackage{times}
\usepackage{latexsym}

\usepackage[T1]{fontenc}

\usepackage{algorithm}
\usepackage{algorithmic}
\usepackage{amsmath}
\usepackage{arydshln}
\usepackage{tabularray}
\usepackage{multirow}
\usepackage{graphicx}
\usepackage{xcolor}

\usepackage{amssymb}
\usepackage{pifont}
\newcommand{\cmark}{\ding{51}}%
\newcommand{\xmark}{\ding{55}}%
\DeclareMathOperator*{\argmin}{arg\,min}
\DeclareMathOperator*{\argmax}{arg\,max}
\usepackage{hyperref}
\usepackage{newfloat}
\usepackage{listings}

\usepackage[utf8]{inputenc}

\usepackage{microtype}

\usepackage{inconsolata}

\usepackage{graphicx}
\usepackage{booktabs}

\definecolor{darkgreen}{rgb}{0.0, 0.5, 0.0}

\title{\textit{Learning Together to Perform Better}:\ Teaching Small-Scale LLMs to Collaborate via Preferential Rationale Tuning}

\author{
  \textbf{Sohan Patnaik}\footnotemark[1],
  \textbf{Milan Aggarwal}\footnotemark[1],
  \textbf{Sumit Bhatia},
  \textbf{Balaji Krishnamurthy}
  \\
  Media and Data Science Research (MDSR) Lab, Adobe
  \\
  \small{\textbf{Correspondence:} \href{mailto:soha@adobe.com}{soha@adobe.com}, \href{mailto:milaggar.adobe.com}{milaggar@adobe.com}}
}

\newcommand{\approachName}{COLLATE}

\begin{document}
\maketitle
\begin{abstract}
LLMs such as GPT-4 have shown a remarkable ability to solve complex questions by generating step-by-step rationales. Prior works have utilized this capability to improve smaller and cheaper LMs (say, with 7B parameters). However, various practical constraints, such as copyright and legal issues, owing to lack of transparency in the pre-training data of large (often closed) models, prevent their use in commercial settings. Little focus has been given to improving the \textit{innate} reasoning ability of smaller models without distilling information from larger LLMs. To address this, we propose \approachName, a trainable framework that tunes a (small) LLM to generate those outputs from a pool of diverse rationales that selectively improves the downstream task. \approachName\ enforces multiple instances of the \textit{same} LLM to exhibit distinct behavior and employs them to generate rationales to obtain diverse outputs. The LLM is then tuned via preference optimization to choose the candidate rationale which maximizes the likelihood of ground-truth answer. \approachName\ outperforms several trainable and prompting baselines on 5 datasets across 3 domains - maths problem solving, natural language inference, and commonsense reasoning. We show the efficacy of \approachName\ on LLMs from different model families across varying parameter scales (1B to 8B) and demonstrate the benefit of multiple rationale providers guided by the end task through ablations. Code is released \href{https://github.com/Sohanpatnaik106/collate}{here}.
\end{abstract}

\section{Introduction}
\renewcommand\thefootnote{*}
\footnotetext{Equal contribution.}

Large Language Models (LLMs) are ubiquitously used to solve various Natural Language Processing (NLP) tasks~\citep{brown2020language}. Due to cost and latency constraints, small-scale LLMs are more suitable for deployment in applications. Given a task-specific dataset consisting of input-output pairs, the LLM is trained on this data through Supervised Fine-Tuning (SFT) to perform the task. Additionally, prior methods have shown that soliciting the LLM to first elucidate a rationale describing the steps required to derive the answer yields better performance compared to generating the final response directly~\citep{wei2022emergent, NEURIPS2022_8bb0d291}. Such a capability is notably exhibited by massive-scale LLMs such as GPT-4~\citep{achiam2023gpt}, PaLM-540B~\citep{chowdhery2022palmscalinglanguagemodeling} etc. which solve complex questions by generating step-by-step rationales~\citep{NEURIPS2022_9d560961}. Hence, several works improve smaller LLMs through larger LLMs by using their generated rationales~\citep{hsieh-etal-2023-distilling, tunstall2023zephyr}.

However, lack of transparency in the pre-training data of larger (often closed) LMs prevents their use in commercial settings due to legal constraints. Models like GPT-4 restrict using their outputs to train other models for commercial use. Hence, the responses of such models can only be used through prompting. Further, long-term use of such LLMs and their APIs can be costly, compute-intensive, and unreliable from a maintenance perspective. Limited attention has been given to developing reasoning ability of small-scale efficient LLMs without relying on any other external LLM. Although Chain-of-Thought (CoT) dataset~\citep{kim2023cot} (comprising of instruction-rationale-response triples) was introduced to teach small LMs to generate rationales via Instruction Fine-Tuning (IFT), it does not contain specific rationale annotations needed for each task. Likewise, prior works such as SPIN~\citep{chen2024self} proposes to tune LLM via Direct Preference Optimisation (DPO - please refer appendix \ref{sec:app_dpo})~\citep{NEURIPS2023_a85b405e} by selecting ground-truth (GT) response as the chosen output over LLM-generated answer. However, lack of availability of annotated rationales for an end-task limits their applicability. Other methods~\citep{yuan2024selfrewardinglanguagemodels} rely on the scale of large LMs to generate diverse responses and rate them via LLM-as-a-judge paradigm~\citep{NEURIPS2023_91f18a12} for preference tuning. Such approaches do not generalise well to small-scale LLMs as demonstrated in the experiments section.



Given these constraints, we ask an important research question - \textit{``Can we teach a small-scale LLM to interact with itself without relying on any other LLM to refine the ability of generating rationales which improves a given end-task?"} We hypothesize that LLMs acquire adequate knowledge during pre-training that is required to generate reasonable rationales. However, the output distribution of the rationales generated by the LLM has to be aligned in a manner that is guided by the final task. Driven by this intuition, we generate diverse rationales for each task sample and optimise the LLM to selectively choose the rationale which enhances the chances of generating correct answer. To achieve this, we propose \approachName\ (Tea\underline{\textbf{C}}hing to C\underline{\textbf{OLL}}aborate via Preferenti\underline{\textbf{A}}l Ra\underline{\textbf{T}}ional\underline{\textbf{E}} Tuning), a framework that leverages multiple \textit{Rationale Provider} LLMs to obtain the set of diverse rationales. The rationale providers are obtained by creating multiple instances of the same LLM such that they exhibit distinct behavior. Given a task instruction, the rationale providers are each employed to generate rationale to create a pool of rationales. Each output in the obtained set of rationales is then assigned a \textit{usefulness score} -- estimated as the likelihood of generating ground-truth answer for the given task conditioned on rationale in input. The usefulness score is used to rank and select the rationales to tune the LLM through DPO. We pick only those samples for DPO where best rationale enhances ground-truth answer likelihood compared to not using any rationale.


We evaluate the effectiveness of \approachName\ based on the utility of the generated rationales to improve performance of small-scale LLMs on five datasets across three diverse task domains: 1) Maths Word Problem Solving - GSM8K~\citep{cobbe2021training}, 2) Natural Language Inference - PIQA~\citep{Bisk_Zellers_Le_bras_Gao_Choi_2020} and WinoGrande~\citep{Sakaguchi_Le_Bras_Bhagavatula_Choi_2020}, and 3) Commonsense Reasoning - CSQA~\citep{talmor-etal-2019-commonsenseqa} and HellaSwag~\citep{zellers-etal-2019-hellaswag}. We observe that \approachName\ outperforms several baselines (\S~\ref{subsec:baselines_comparison}) by gains of up to $7\%$ without dependence on other larger LLMs (unlike prior methods) for generating diverse rationales and rating their quality. Further, we show the efficacy of \approachName\ on improving a variety of LLMs across varying parameter-scales of 1B to 8B (\S~\ref{subsec:scale_exp}). We also conduct \textbf{human studies} which show that the rationales from \approachName{} are reliable, helpful and \textbf{diverse} (\S~\ref{sec:human_study}). Further, we conduct extensive ablations to validate - \textbf{1)} employing rationale providers for diverse rationales as opposed to sampling-based decoding; and \textbf{2)} using end-task guided likelihood-based rationale selection and sample filtration for DPO (Appendix~\ref{app:abl_studies}).


\section{Related Work}

\textbf{Prompt-based Reasoning Generation:} It has been shown that generating intermediate reasoning chains improves the performance of large-scale LLMs~\citep{wei2022emergent, NEURIPS2022_8bb0d291}. Chain-of-Thought (CoT) prompting~\citep{NEURIPS2022_9d560961} is the first such technique which performs this by showing exemplar demonstrations of step-by-step reasoning in the prompt. Consequently, some other methods have focused on improving the diversity of the exemplar demonstrations in the prompt~\citep{zhang2023automatic, diao2024active, li2023finding, li2023mot}. Tree-of-Thought (ToT)~\citep{NEURIPS2023_271db992} was introduced as a generalisation of CoT where different reasoning paths are organised in the form of a tree such that the LLM can look-ahead or backtrack in tree to yield the optimal reasoning chains. Further, many self-correction methods were proposed where the LLM identifies and rectifies its own mistakes~\citep{NEURIPS2023_91edff07, wang2023boosting}. However, it was shown that a single LLM struggles to improve its rationales through just prompting-based mechanisms in the absence of external feedback~\citep{huang2024large, valmeekam2023can, stechly2023gpt}.



\begin{figure*}[t] 
    \centering 
    \includegraphics[width=\textwidth]{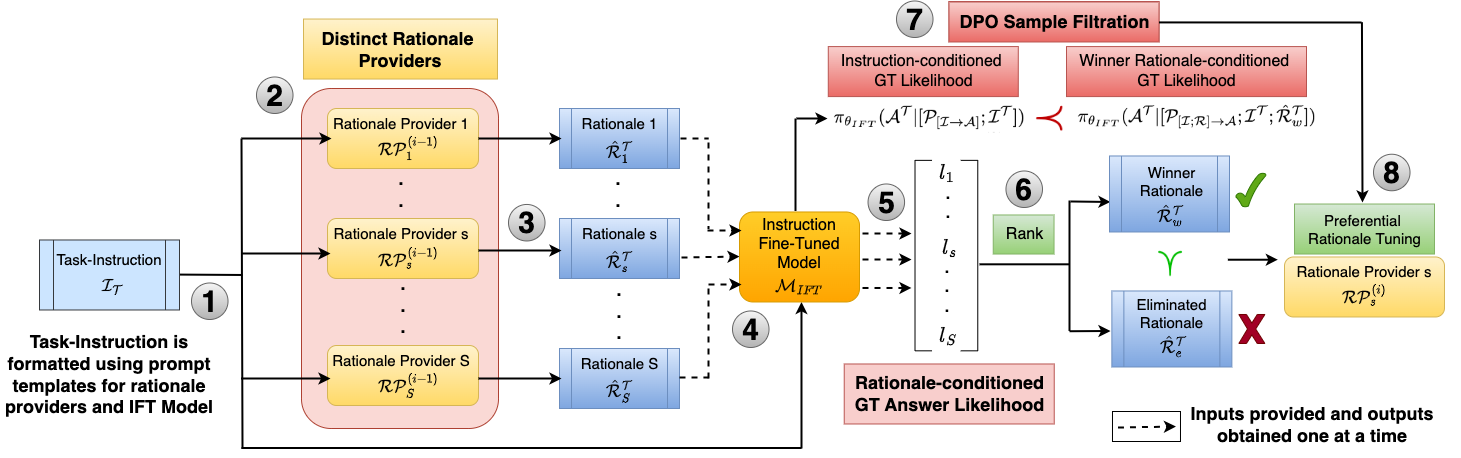} 
    \caption{Overview of architecture of \approachName. The task-instruction ($\mathcal{I}^{\mathcal{T}}$) is pre-processed (step 1) and given as input to distinct rationale providers - $\mathcal{RP}^{(i-1)}_{s}$ (step 2) to obtain set of diverse rationales - $\{\hat{\mathcal{R}}_{s}^{\mathcal{T}}\}$ (step 3). IFT version of the LLM - $\mathcal{M_{IFT}}$ is used to estimate likelihood ($l_s$) of generating ground-truth (GT) answer conditioned on task-instruction and each rationale (steps 4 and 5) which is used to rank rationales to identify winning ($\hat{\mathcal{R}}_{w}^{\mathcal{T}}$) and eliminated ($\hat{\mathcal{R}}_{e}^{\mathcal{T}}$) rationales (step 6). If the winning-rationale enhances likelihood of obtaining GT answer compared to using just the instruction (step 7), the sample is used to perform DPO to obtain rationale provider $\mathcal{RP}^{(i)}_{s}$ (step 8).}
    \label{fig:collate_arch_diag} 
\end{figure*}

\noindent \textbf{Rationale Enhancement via Preference Training:} Some works optimise reasoning chains by obtaining a set of rationales and applying Direct Preference Optimisation (DPO)~\citep{NEURIPS2023_a85b405e}. SPIN~\citep{chen2024self} proposes to choose the final ground-truth (GT) answer over the LLM generated final response for DPO training. Likewise, \citet{lai2024step} shows that DPO can be applied at each step of the reasoning chain which requires extensive annotations. Lack of availability of GT rationales on a new task limit their applicability. On the other hand, we employ different instances of the same LLM to generate diverse rationales for selection during DPO. \citet{zhang2024chain} proposed an alternative way to obtain multiple rationales by extracting reasoning chains sampled at each step of ToT tree search. Very recently, \citet{wang2024selftrainingdirectpreferenceoptimization} proposed sampling-based decoding to obtain diverse rationales for math problems such that rationales which contain GT answer are considered as preferred outputs. Such an approach cannot be extended to non-math domains where just the presence of answer is not indicative of rationale quality. \citet{yuan2024selfrewardinglanguagemodels} showed that very-large LMs (Llama-2 70B) can be used to rank outputs through prompting. Contrastingly, \approachName\ rates rationale quality by leveraging LLM's likelihood of GT answer conditioned on the rationale. \\
\textbf{Response Synthesis through Multi-LLM Interaction:} Several works have explored ensembling multiple LLMs to improve overall response quality~\citep{jiang-etal-2023-llm, yu2024explanationaware, juneja-etal-2023-small, ulmer2024bootstrapping, lu2024blending}. Zephyr~\citep{tunstall2023zephyr} distills responses generated by Falcon~\citep{penedo2023refinedwebdatasetfalconllm}, Llama~\citep{touvron2023llama2openfoundation} etc. into Mistral-7B~\citep{jiang2023mistral7b} by rating them using GPT-4~\citep{achiam2023gpt} to perform DPO. Likewise, \citet{hsieh-etal-2023-distilling} distills rationales from PaLM-540B model~\citep{chowdhery2022palmscalinglanguagemodeling} into T5~\citep{raffel2023exploringlimitstransferlearning} through SFT. Few approaches personalise the feedback from larger teacher LLM based on weaknesses of smaller student LLM~\citep{wang-li-2023-learning, saha2023can, jiang-etal-2023-lion}. \citet{kang2023knowledgeaugmented} aim at improving LLMs in a task-oriented manner by retrieving knowledge using rationales obtained from a larger LLM. Orca~\citep{mukherjee2023orcaprogressivelearningcomplex} argues that IFT over GPT-generated outputs makes smaller LLM imitate just the style~\citep{gudibande2024the} and proposes to train smaller LLM on explanation traces of GPT-4. Lack of transparency in training data of larger LLMs hinders their commercial use due to legal implications. Other methods mimic the way humans conduct discussions~\citep{mousavi2023n} by facilitating prompt-based deliberation ~\citep{yin-etal-2023-exchange}. We show that such methods work well only with large LMs but performs poorly with smaller LMs.


\section{Methodology}



\textbf{Overview of our Approach}: Figure~\ref{fig:collate_arch_diag} provides an outline of the proposed \approachName{} framework to generate better rationales for improving end-task performance \textit{without} relying on distilling information from any external LLM. To achieve this, we first train the instruction fine-tuned (IFT) version of the base LLM - $\mathcal{M}_{IFT}$ (\S~\ref{subsec:ift}). We then create multiple rationale providers by cloning $S$ instances of $\mathcal{M}_{IFT}$ and optimise them on separate data splits to ensure they exhibit distinct behaviour (\S~\ref{subsec:dis_rat_pro}). The rationale providers are tuned through preference optimisation (DPO) to selectively generate those outputs from a pool of diverse rationales which help in improving the end-task (\S~\ref{sec:task_guided_dpo}). The set of diverse rationales is obtained by generating reasoning chains from the \textit{Rationale-Provider} ($\mathcal{RP}$) LLMs (steps 1-3 in fig.\ref{fig:collate_arch_diag}). Once the candidate set of rationales is obtained, a usefulness score is assigned to each rationale based on end-task performance. The score is estimated as the likelihood of generating the ground-truth (GT) answer by $\mathcal{M}_{IFT}$ conditioned on the rationale in input (steps 4-5 in fig.\ref{fig:collate_arch_diag}). The score is used to select the winning and eliminated outputs for DPO (steps 6-8 in fig.\ref{fig:collate_arch_diag}). We feed only high-quality samples for DPO based on whether the winning rationale enhances likelihood of generating GT compared not using any rationale (step 7 in fig.\ref{fig:collate_arch_diag}). We now describe the individual components of \approachName\ in detail.


\subsection{Multi-Mode Instruction Fine-Tuning}
\label{subsec:ift}
Conventional Instruction Fine-Tuning (IFT) aims at enabling the LLM $\mathcal{M}$ to follow a given instruction to generate the final response accordingly. However, we require the instruction-tuned model $\mathcal{M}_{IFT}$ for two additional objectives - 1) generate rationale required to derive the final answer given the instruction as input; 2) generate/estimate the likelihood of producing an answer given the instruction and rationale as input. To enable $\mathcal{M}_{IFT}$ to operate in different modes, we format the samples using different prompts (please refer appendix \ref{sec:app_prompt_template}) during training to indicate the model about the mode in which it needs to generate the output. We leverage a dataset $\mathcal{D}^{rationale}_{IFT}$ which comprises of samples containing rationales in addition to instruction-answer pairs to enable the two additional modes discussed above. Formally, given an instruction $\mathcal{I}$, rationale $\mathcal{R}$ and final answer $\mathcal{A}$, we perform IFT in three modes using cross-entropy loss through teacher forcing~\citep{NIPS2017_3f5ee243}:
\vspace{-2mm}

\small
\begin{align}
    \mathcal{L}_{\mathcal{I} \rightarrow \mathcal{A}} &= -log\ p(\mathcal{A}_t|\ [\mathcal{P}_{\mathcal{I} \rightarrow \mathcal{A}}; \mathcal{I}; \mathcal{A}_{<t}], \theta_{IFT}) \\
    \mathcal{L}_{\mathcal{I} \rightarrow \mathcal{R}} &= -log\ p(\mathcal{R}_t|\ [\mathcal{P}_{\mathcal{I} \rightarrow \mathcal{R}}; \mathcal{I}; \mathcal{R}_{<t}], \theta_{IFT}) \\
    \mathcal{L}_{[\mathcal{I}; \mathcal{R}] \rightarrow \mathcal{A}} &= -log\ p(\mathcal{A}_t|\ [\mathcal{P}_{[\mathcal{I}; \mathcal{R}] \rightarrow \mathcal{A}}; \mathcal{I}; \mathcal{R}; \mathcal{A}_{<t}], \theta_{IFT})
\end{align}


\normalsize
\noindent where, $p$ represents probability, $\mathcal{A}_t$ and $\mathcal{R}_t$ depict the $t^{th}$ token in ground-truth answer and rationale respectively, $<t$ indicates tokens before $t^{th}$ index, $\mathcal{P}_m$ and $\mathcal{L}_m$ are the prompt format and loss function respectively for the $m^{th}$ mode; $[;]$ represents the formatting operation to prepare LLM input after arranging the instruction, answer and/or rationale into mode-specific prompt $\mathcal{P}_m$, and $\theta_{IFT}$ is the set of trainable LLM parameters. For samples in IFT data which do not contain rationales, only loss $\mathcal{L}_{\mathcal{I} \rightarrow \mathcal{A}}$ is applied. Once $\mathcal{M}_{IFT}$ is trained, it is used to obtain distinct rationale provider LLMs and rank rationales generated by them for task-guided preferential rationale tuning in subsections to follow.



\subsection{Distinct Rationale Providers}
\label{subsec:dis_rat_pro}
To improve rationale generation ability on a new task, \approachName\ generates diverse rationales to obtain a candidate set of outputs that can be used to optimise the LLM via DPO. To obtain diverse rationales, we train multiple Rationale Provider (RP) LLMs - $\mathcal{RP}_1$, $\mathcal{RP}_2$, ..., $\mathcal{RP}_S$ such that they can be leveraged to obtain a diverse set of outputs. The rationale providers are obtained by cloning $S$ instances of $\mathcal{M}_{IFT}$ and tuning them for rationale generation on different data splits so that they exhibit different behavior. Specifically, we consider a dataset containing rationale annotations - $\mathcal{D}^{rationale}$ and divide it into $S$ equal splits randomly. We denote $s^{th}$ data split as $\mathcal{D}_s^{rationale} = \{d_{1}^{s}, ..., d_{N/S}^{s}\}$, where $1\leq s \leq S$, $N$ is the number of samples in $\mathcal{D}^{rationale}$, and $d_{j}^{s}$ denotes $j^{th}$ sample in split $s$ comprising of instruction-rationale-response triples - $\{\mathcal{I}_{j}^{s}, \mathcal{R}_{j}^{s}, \mathcal{A}_{j}^{s} \}$. We assign the $s^{th}$ data split to rationale-provider $\mathcal{RP}_{s}$ and prompt it to generate the rationale as:
\vspace{-1mm}
\begin{align}
    \hat{\mathcal{R}}_{j}^{s} = \mathcal{RP}_s([\mathcal{P}_{\mathcal{I}\rightarrow\mathcal{\mathcal{R}}};\mathcal{I}_{j}^{s}]\ |\ \theta_{RP_s})
\end{align}

\noindent The rationale provider $\mathcal{RP}_s$ is then fine-tuned through DPO by choosing the ground-truth rationale $\mathcal{R}_{j}^{s}$ as the preferred option over the generated rationale $\hat{\mathcal{R}}_{j}^{s}$. Formally, $\mathcal{RP}_s$ is optimised using the following loss function $\mathcal{L}_{\mathcal{RP}_s}$:

\small
\begin{align}
    \mathcal{L}_{\mathcal{RP}_s} = -log\sigma[\beta(log\frac{\pi_{\mathcal{RP}_s}(\mathcal{R}_{j}^{s})}{\pi_{\mathcal{M}_{IFT}}(\mathcal{R}_{j}^{s})} - log\frac{\pi_{\mathcal{RP}_s}(\hat{\mathcal{R}}_{j}^{s})}{\pi_{\mathcal{M}_{IFT}}(\hat{\mathcal{R}}_{j}^{s})})]
\end{align}

\normalsize
where, $\pi_{\mathcal{Z}}(y)$ represents the likelihood of generating tokens in $y \in \{\mathcal{R}_{j}^{s}, \hat{\mathcal{R}}_{j}^{s}\}$ by model $\mathcal{Z} \in \{\mathcal{RP}_s, \mathcal{M}_{IFT} \}$, $\sigma$ is sigmoid activation, $\beta$ (= 0.1) is a coefficient to control deviation from reference model $\mathcal{M}_{IFT}$. This yields a set of $S$ distinct rationale providers which are used to generate diverse rationales for a given task-specific instruction. The generated rationales are used to optimise the rationale providers further in a task-guided manner as described in next sub-section. We refer to $s^{th}$ rationale provider obtained till this stage as $\mathcal{RP}_{s}^{(0)}$ as a convention for subsequent discussions.




\subsection{Task-Based Preferential Rationale Tuning}
\label{sec:task_guided_dpo}
For a given end-task $\mathcal{T}$, we denote the corresponding dataset as $\mathcal{D}^{\mathcal{T}}$ comprising of instruction-answer pairs of the form $(\mathcal{I}^{\mathcal{T}}, \mathcal{A}^{\mathcal{T}})$. However, the supervised dataset for the given task does not contain rationale annotations. To mitigate this, we leverage distinct rationale providers to construct a set of diverse rationales. Quality of a rationale is determined based on its usefulness to enhance end-task performance i.e. likelihood of generating the ground-truth answer $\mathcal{A}^{\mathcal{T}}$. Each rationale provider is then optimised iteratively to prefer generating better rationale candidate as opposed to other options which result in lower likelihood of generating the ground-truth answer. The rationale providers are tuned for numerous iterations by conducting multiple passes over $\mathcal{D}^{\mathcal{T}}$.


In particular, consider the $i^{th}$ iteration ($i \in \{1, 2\}$) such that the task-instruction $\mathcal{I}^{\mathcal{T}}$ is given as input to each rationale provider - $\mathcal{RP}_{s}^{(i-1)}$ (obtained till last $(i-1)^{th}$ iteration) separately to generate the corresponding rationale as depicted in the following equation (steps 1-3 in fig.\ref{fig:collate_arch_diag}):
\vspace{-0.5mm}
\begin{align}
    \hat{\mathcal{R}}^{\mathcal{T}}_{s} = \mathcal{RP}_{s}^{(i-1)}([\mathcal{P}_{\mathcal{I}\rightarrow\mathcal{\mathcal{R}}};\mathcal{I}^{\mathcal{T}}]\ |\ \theta_{RP^{(i-1)}_{s}})
\end{align}

\noindent We obtain a set of generated rationales - $\{ \hat{\mathcal{R}}^{\mathcal{T}}_{1}, \hat{\mathcal{R}}^{\mathcal{T}}_{2}, ..., \hat{\mathcal{R}}^{\mathcal{T}}_{S} \}$. Each rationale in the set is assigned a usefulness score $l_s$ ($1 \leq s \leq S$) by estimating the likelihood of generating the ground-truth (GT) answer $\mathcal{A}^{\mathcal{T}}$ by the instruction fine-tuned model $\mathcal{M}_{IFT}$ conditioned on rationale and task-instruction as input - $\pi_{\theta_{IFT}}(\mathcal{A}^{\mathcal{T}}\ |\ [\mathcal{P}_{[\mathcal{I};\mathcal{R}]\rightarrow\mathcal{A}}; \mathcal{I}^{\mathcal{T}}; \hat{\mathcal{R}}^{\mathcal{T}}_{s}])$ (steps 4 and 5 in fig.\ref{fig:collate_arch_diag}). The rationales $\hat{\mathcal{R}}^{\mathcal{T}}_{s}$ ($1 \leq s \leq S$) are ranked based on their usefulness score $l_s$ such that the first and last elements in the ranked list are selected as winner ($\hat{\mathcal{R}}^{\mathcal{T}}_{w}$) and eliminated ($\hat{\mathcal{R}}^{\mathcal{T}}_{e}$) rationales respectively (step 6 in fig.~\ref{fig:collate_arch_diag}):
\vspace{-0.5mm}
\begin{align}
    w &= \argmax_{1 \leq s \leq S} \{l_s\}\ ;\ e = \argmin_{1 \leq s \leq S} \{l_s\}\
\end{align}

\noindent To maintain the distinctness in the behaviour of different rationale providers, we tune them on different splits of $\mathcal{D}^{\mathcal{T}}$. The task-specific dataset $\mathcal{D}^{\mathcal{T}}$ is divided into $S$ equal splits randomly such that split $\mathcal{D}^{\mathcal{T}}_{s}$ ($1 \leq s \leq S$) is assigned to optimise rationale provider $\mathcal{RP}^{(i)}_{s}$. Without loss of generality, consider ($\mathcal{I}^{\mathcal{T}}, \mathcal{A}^{\mathcal{T}}) \in \mathcal{D}^{\mathcal{T}}_{s}$. $\hat{\mathcal{R}}_{w}^{\mathcal{T}}$ and $\hat{\mathcal{R}}_{e}^{\mathcal{T}}$ are used to tune $s^{th}$ rationale provider $\mathcal{RP}^{(i)}_{s}$ in current iteration $i$ via preference optimisation (step 8 in fig.~\ref{fig:collate_arch_diag}) using the loss $\mathcal{L}_{\mathcal{RP}^{(i)}_s}^{\mathcal{T}}$:

\small
\begin{align}
    \mathcal{L}_{\mathcal{RP}^{(i)}_s}^{\mathcal{T}} = -log\sigma[\beta(log\frac{\pi_{\mathcal{RP}^{(i)}_s}(\hat{\mathcal{R}}_{w}^{\mathcal{T}})}{\pi_{\mathcal{RP}_{s}^{(j)}}(\hat{\mathcal{R}}_{w}^{\mathcal{T}})} - log\frac{\pi_{\mathcal{RP}^{(i)}_s}(\hat{\mathcal{R}}_{e}^{\mathcal{T}})}{\pi_{\mathcal{RP}_{s}^{(j)}}(\hat{\mathcal{R}}_{e}^{\mathcal{T}})})]
\end{align}

\normalsize
\noindent where, $\beta=0.1$ and $j=(i-1)$. It can be noted that the previous iteration $(i-1)$ version of the rationale provider - $\mathcal{RP}^{(i-1)}_{s}$ is used as the reference model for the DPO training to obtain $\mathcal{RP}^{(i)}_{s}$.

\noindent \textbf{Likelihood-based Sample Filtration}: To ensure that high-quality samples are used to tune rationale providers, we apply a filtration criteria to remove the samples where the winning rationale does not enhance the likelihood of generating the GT. Formally, we retain the samples $(\mathcal{I}^{\mathcal{T}}, \mathcal{A}^{\mathcal{T}})$ which qualify the following criteria (step 7 in fig.\ref{fig:collate_arch_diag}):

\small
\begin{align}
    \pi_{\theta_{IFT}}(\mathcal{A}^{\mathcal{T}}|[\mathcal{P}_{[\mathcal{I};\mathcal{R}]\rightarrow\mathcal{A}}; \mathcal{I}^{\mathcal{T}}; \hat{\mathcal{R}}^{\mathcal{T}}_{w}]) >
    \pi_{\theta_{IFT}}(\mathcal{A}^{\mathcal{T}}|[\mathcal{P}_{[\mathcal{I}\rightarrow\mathcal{A}]}; \mathcal{I}^{\mathcal{T}}])
    \label{eq:like-filtering}
\end{align}

\normalsize
\noindent Equation~\ref{eq:like-filtering} compares the likelihood of generating the ground-truth by $\mathcal{M}_{IFT}$ for the task instruction in the absence and presence of the winning rationale in the input. The sample is used for DPO training if the winning rationale enhances the likelihood compared to not using any rationale.

\begin{table*}[t]
    \centering
    \resizebox{0.95\textwidth}{!}{%
    \begin{tabular}{lccccc}
        \toprule
        \multirow{2}{*}{\textbf{Method}} & \multicolumn{1}{c}{\textbf{Maths}}  & \multicolumn{2}{c}{\textbf{NLI}} & \multicolumn{2}{c}{\textbf{Comonsense}} \\
        & \textbf{GSM8K} & \textbf{WinoGrande} & \textbf{PIQA} & \textbf{HellaSwag} & \textbf{CSQA}  \\ \midrule
        \textbf{\textit{Prompt-based Rationale Refinement}} & & & & & \\
        Chain-of-Thought~\citep{NEURIPS2022_9d560961} & 13.55 & 28.19 & 28.81  & 55.89 & 36.27 \\
        CoT Self-Consistency~\citep{wang2023selfconsistency} & 14.56 & 33.32 & 31.04 & 51.23 & 37.17\\
        Tree-of-Thought~\citep{NEURIPS2023_271db992} & 17.31 & 30.10 & 31.61 & 63.33 & 39.49  \\
        Exchange-of-Thought~\citep{yin-etal-2023-exchange} & 26.80 & 33.29 & 38.17 & 72.13 & 55.31  \\
        \midrule
        \textbf{\textit{Task-specific Supervised Fine-Tuning}} & & & & & \\
        open-llama-7b-v2 SFT~\citep{openlm2023openllama} & 30.76 & 48.81 & 50.94 & 94.79 & 73.82  \\
        Distilling Step-by-Step~\citep{hsieh-etal-2023-distilling} & 38.41 & 50.69 & 50.65 & 96.01 & 75.02 \\
        \midrule
        
        \textbf{\textit{Training-driven Rationale Enhancement}} & & & & & \\
        Self-Rewarding LMs~\citep{yuan2024selfrewardinglanguagemodels} & 39.13 & 41.44 & 47.41 & 93.26 & 71.77  \\
        SPIN~\citep{chen2024self} & 48.41 & 50.58 & 51.94 & 98.14 & 78.76  \\
        \midrule
        \textbf{\approachName\ (ours)}  & \textbf{53.48} & \textbf{53.49} & \textbf{59.57} & \textbf{99.21} & \textbf{80.13}  \\
        \bottomrule
    \end{tabular}%
    }
    \caption{Comparison of \approachName\ with multiple categories of baselines on five datasets over three task domains - Maths Problem Solving, Natural Language Inference (NLI) and Commonsense Reasoning. \approachName\ performs better than baselines. We use common LLM backbone i.e. open-llama-v2-7b for all methods in this table.} 
    \label{tab:baseline_comparison}
\end{table*}

\section{Experiments and Evaluation}


\noindent \textbf{Datasets:} 
We evaluate \approachName\ on three diverse reasoning task domains using five datasets - \textbf{1) Maths Problem Solving} on GSM8K~\citep{cobbe2021training}; \textbf{2) Natural Language Inference (NLI)} using PIQA~\citep{Bisk_Zellers_Le_bras_Gao_Choi_2020} and WinoGrande~\citep{Sakaguchi_Le_Bras_Bhagavatula_Choi_2020}; and \textbf{3) Commonsense Reasoning} through CSQA~\citep{talmor-etal-2019-commonsenseqa} and HellaSwag~\citep{zellers-etal-2019-hellaswag}. \textbf{GSM8K} comprises of maths word problems requiring understanding of problem and sequence of calculations. \textbf{PIQA} requires understanding of physical relation between objects and comprises of samples with a goal text coupled with two candidate texts out of which only one can lead to the goal. \textbf{WinoGrande} is a very challenging co-reference resolution task, comprising of a statement with two parts such that the latter half refers to some entity in the first part. \textbf{CSQA} tests model's ability to answer MCQ questions by picking correct choice using commonsense knowledge. \textbf{HellaSwag} evaluates ability to predict continuation of a context by choosing most plausible ending. Number of samples in train/test splits of each dataset is GSM8k - 7.5k/1.3k, PIQA - 16k/3k, WinoGrande - 40k/1.7k, CSQA - 9.7k/1.2k and HellaSwag - 39.9k/10k. Please refer to the appendix \ref{sec:app_dataset_samples} for examples from each dataset.

\noindent \textbf{Implementation Details:} In all training runs, we use a batch size (BS) of 16 on 8 80GB A100 GPUs (BS of 2/GPU), a learning rate of 1e-5, bfloat16 precision with cosine annealing~\citep{loshchilov2017sgdr} using AdamW optimizer~\citep{loshchilov2018decoupled}. We leverage DeepSpeed Zero 2 with sharding of optimizer states and gradients across GPUs and enable gradient check-pointing.

\noindent \textbf{1) Multi-Mode Instruction Fine-Tuning (IFT) -} We experiment with different LLM backbones $\mathcal{M}$ (open-llama-v2-7b, OLMo-1B, Phi3-3.8B, Qwen1.5-4B and LLaMA3-8B) to obtain $\mathcal{M}_{IFT}$. We use a random subset of 140k samples from CoT-Collection~\citep{kim2023cot} as $\mathcal{D}^{rationale}_{IFT}$ to enable two additional IFT modes - ${\mathcal{I} \rightarrow \mathcal{R}}$ and $[\mathcal{I;R}] \rightarrow \mathcal{A}$. We use 40k samples of Dolly-HHRLHF~\citep{mosaicml2023dolly_hhrlhf} and Open Assistant datasets combined to create samples for $\mathcal{I} \rightarrow \mathcal{A}$ mode. The IFT training is performed for 2 epochs.

\noindent \textbf{2) Distinct Rationale Providers -} We tune $S=2$ clones of $\mathcal{M}_{IFT}$ using subset of 195k samples in CoT-Collection ($\mathcal{D}^{rationale}$) to obtain rationale providers. DPO training is performed for 5 epochs.

\noindent \textbf{3) Task-guided Preferential Rationale Tuning -} We carry out 2 iterations of task-guided DPO with 10 epochs in each iteration. The rationale providers are evaluated on a val-split of CoT comprising of 8k samples to identify the rationale provider to be used for evaluation as described next. 

\noindent \textbf{Evaluation Procedure:} We evaluate \approachName\ and baselines by conducting rationale-conditioned Supervised Fine-Tuning (SFT) of the base LLM $\mathcal{M}$ (for 3 epochs) on task-specific dataset comprising of pairs of input instruction and final answer. The rationale generated by a method is given as additional input along with task instruction during this SFT. The accuracy achieved is indicative of usefulness of rationales. We use a single rationale provider while assessing rationales generated by \approachName. The best rationale provider is selected based on likelihood estimation on CoT val-split samples where we measure probability of generating ground-truth answer by the IFT model $\mathcal{M}_{IFT}$ conditioned on the rationale. The rationale provider that yields maximum value (averaged across samples) is used for evaluation. Figure~\ref{fig:corr_like_acc} in Appendix~\ref{app:rp_selection} shows correlation plot between task test-accuracy and likelihood indicating that rationale provider with best likelihood is a suitable choice.

\begin{table*}[ht]
    \centering
    \resizebox{0.95\textwidth}{!}{%
    \begin{tabular}{p{200mm}}
        \toprule
         \textbf{Task-Instruction and Rationales Generated by Best Baseline (SPIN) and Rationale Providers - $\mathcal{RP}_1$ and $\mathcal{RP}_2$ in \approachName} \\ \midrule
         \textbf{Instruction:} Betty is saving money for a new wallet which costs \$100. Betty has only half of the money she needs. Her parents decided to give her \$15 for that purpose, and her grandparents twice as much as her parents. How much more money does Betty need to buy the wallet?\\ 
        \textbf{Baseline:} Betty has \$50. Her parents give her \$15, and her grandparents give her twice that, which is \$100. So, she gets an additional \$115. Now she has \$50 + \$115 = \$165. Therefore, she needs \$65 more to buy the wallet. \textcolor{red}{\xmark}\\
         \textbf{Rationale Provider 1:} Betty needs \$100 for a new wallet. She currently has half of that amount, i.e. \$100/2=\$50. Her parents contribute \$15 towards her goal, and her grandparents contribute twice as much as her parents, which is \$30. Betty currently has \$50 + \$15 + \$30 = \$95. We need to subtract her current amount from the price of the wallet, which is \$100 - \$95 = \$5. \textcolor{teal}{\cmark}\\ 
         \textbf{Rationale Provider 2:} Betty needs \$100 to buy the wallet. She has \$50. Her parents give her \$15, and her grandparents give her twice as much, which is \$30. The amount she needs is \$55. \textcolor{red}{\xmark}\\
        \bottomrule
    \end{tabular}%
    }
    \caption{Qualitative comparison of rationales generated by best baseline (SPIN) and rationale providers in \approachName\ on GSM8K. \approachName\ generates better rationales by virtue of employing distinct rationale providers.}
    \label{tab:qualitative_samples_main_paper}
\end{table*}


\subsection{Comparison of \approachName\ with Baselines}
\label{subsec:baselines_comparison}
We present a detailed comparative analysis of \approachName\ with three categories of baselines (Table~\ref{tab:baseline_comparison}) - \textbf{(i) Prompting} techniques which enable the LLM to either generate reasoning chains in different ways such as Chain-of-Thought~\citep{NEURIPS2022_9d560961}, Tree-of-Thought~\citep{NEURIPS2023_271db992}, etc. or facilitate exchange of feedback between multiple LLMs - Exchange-of-Thought~\citep{yin-etal-2023-exchange}; \textbf{(ii) Task-specific Supervised Fine-Tuning (SFT)} of LLM without any rationales in the input and with rationales generated by the IFT version of the LLM $\mathcal{M}_{IFT}$ - Distilling Step-by-Step~\citep{hsieh-etal-2023-distilling}; and \textbf{(iii) Training-driven Rationale Enhancement} where the LLM is tuned to improve rationales iteratively through DPO - SPIN~\citep{chen2024self} and Self-rewarding Language Models~\citep{yuan2024selfrewardinglanguagemodels}). We employ a common LLM backbone (open-llama-v2-7b) for \approachName\ as well as baselines for a fair comparison.

Table~\ref{tab:baseline_comparison} summarises the results where it can be seen that \approachName\ outperforms all the baselines uniformly across the three task domains. Specifically, it performs better than the best baseline (SPIN) by $\sim5\%$ on GSM8K indicating the helpfulness of rationales from \approachName\ for solving maths problems. Further, \approachName\ achieves significantly better results than SPIN for NLI with an improvement of $\sim7.5\%$ on PIQA and $\sim3\%$ on WinoGrande where most baselines achieve close to random-selection accuracy. Finally, for commonsense reasoning, even though the accuracy of SPIN ($98.14\%$) is already close to $100$ on HellaSwag, \approachName\ still performs better by achieving $50\%$ of remaining possible improvement. In particular, \approachName\ performs better than SPIN by $1.37\%$ on CSQA and $1.07\%$ on HellaSwag.

\begin{table*}[ht]
    \centering
    \resizebox{0.95\textwidth}{!}{%
    \begin{tabular}{lcccccc}
        \toprule
        \multirow{2}{*}{\textbf{Model}} & \textbf{Parameter} & \textbf{Maths} & \multicolumn{2}{c}{\textbf{NLI}} & \multicolumn{2}{c}{\textbf{Commonsense}}\\
        & \textbf{Scale} & \textbf{GSM8K} & \textbf{WinoGrande} & \textbf{PIQA} & \textbf{HellaSwag} & \textbf{CSQA} \\ \midrule
        OLMo~\citep{groeneveld2024olmo} & 1B & 11.89 & 31.31 & 25.98 & 68.02 & 48.19 \\
        \hspace{5mm} \textbf{w/ \approachName\ (ours)} & 1B & \textbf{16.02} & \textbf{36.01} & \textbf{29.94} & \textbf{73.14} & \textbf{55.17} \\ \midrule 
        Phi3~\citep{abdin2024phi} & 3.8B & 21.08 & 42.04 & 37.72 & 76.09 & 57.46 \\
        \hspace{5mm} \textbf{w/ \approachName\ (ours)} & 3.8B & \textbf{28.29} & \textbf{45.28} & \textbf{42.35} & \textbf{81.67} & \textbf{64.14} \\ \midrule
        Qwen1.5~\citep{qwen} & 4B & 19.17 & 39.66 & 34.57 & 80.03 & 56.21 \\
        \hspace{5mm} \textbf{w/ \approachName\ (ours)} & 4B & \textbf{26.03} & \textbf{43.35} & \textbf{39.22} & \textbf{83.31} & \textbf{61.89} \\ \midrule 
        Open-LLaMA-v2~\citep{openlm2023openllama} & 7B & 38.41 & 50.69 & 50.65 & 96.01 & 75.02 \\
        \hspace{5mm} \textbf{w/ \approachName\ (ours)} & 7B & \textbf{53.48} & \textbf{53.49} & \textbf{59.57} & \textbf{99.21} & \textbf{80.13} \\ \midrule 
        LLaMA3~\citep{dubey2024llama} & 8B & 60.68 & 48.21 & 49.35 & 90.35 & 80.39 \\
        \hspace{5mm} \textbf{w/ \approachName\ (ours)} & 8B & \textbf{69.09} & \textbf{54.41} & \textbf{58.70} & \textbf{92.86} & \textbf{87.42} \\
        \bottomrule
    \end{tabular}%
    }
    \caption{Evaluation of \approachName\ on LLMs of varying scale of parameters (1B to 8B) and different LLMs (OLMo, Phi3, Qwen1.5, Open-llama-v2, Llama3). \approachName\ yields significant gains across scales and models on all tasks.}
    \label{tab:llm_scale}
\end{table*}


Additionally, \approachName\ gives a significant performance boost to the base LLM compared to both, carrying out SFT over task input-output pairs without rationales, and using the rationales generated by the IFT model as additional input during SFT (Distilling Step-by-Step - DSS). Compared to `DSS', there is an increase of $\sim15\%$ on GSM8K, $3.2\%$ on HellaSwag, $\sim5\%$ on CSQA, $\sim 3\%$ on WinoGrande, and $\sim9\%$ on PIQA . Moreover, the baseline `Self-Rewarding Language Models' which leverages larger LLM (Llama-70B) to both generate and rate diverse rationales does not generalise with 7B LM. It performs worse than `DSS' on most tasks. Finally, it is noted that prompting-based methods give significantly lower performance at the scale of 7B LLM. We also perform few-shot prompt-based evaluation (instead of SFT) of rationales from \approachName\ in appendix~\ref{app:few_shot_eval} where we observe a similar trend compared to baselines.

Table~\ref{tab:qualitative_samples_main_paper} compares the quality of rationales generated by rationale providers with best baseline (SPIN). COLLATE generates better rationales by virtue of employing distinct rationale providers. Please refer to appendix~\ref{sec:app_qualitative} for more examples.


\subsection{Does \approachName\ Enhance Small LLMs of Varying Scales and Model Families?}
\label{subsec:scale_exp}
We verify if \approachName\ improves performance of multiple small-scale LLMs of varying scales ranging from \textbf{1B to 8B} parameters belonging to different model families. We compare the accuracy of task-specific SFT version of $\mathcal{M}$ trained using rationales generated by \approachName\ vs. the rationales from IFT model $\mathcal{M}_{IFT}$. Specifically, we experiment with \textbf{1) OLMo-1B}~\citep{groeneveld2024olmo}, \textbf{2) Phi3-3.8B}~\citep{abdin2024phi}, \textbf{3) Qwen1.5-4B}~\citep{qwen}, \textbf{4) open-llama-v2-7B}~\citep{openlm2023openllama}, and \textbf{5) LLaMA3-8B}~\citep{dubey2024llama}. Table~\ref{tab:llm_scale} summarises the results where it can be seen that \approachName\ provides significant performance increase on all the tasks uniformly over parameter-scales and LLM families. Across parameters scale, there is a gain of $4-8\%$ on GSM8K, $2.5-6\%$ on WinoGrande, $4-9\%$ on PIQA, $\sim1.5-5\%$ on HellaSwag and $5-7\%$ on CSQA. Likewise across tasks, there is a boost of $4-7\%$ at 1B scale ; $3-7\%$ increase at $\sim$4B scale; and $\sim1.5-9\%$ gain for 7-8B parameters LLMs.

\subsection{Human Study of \approachName{}'s Rationales}
\label{sec:human_study}
We conducted a human study to evaluate the rationales generated by the rationale providers (details in Appendix~\ref{app:human_study_app}). We asked human evaluators to judge if 1) final rationale obtained from \approachName{} is correct and helpful, and 2) whether the rationale generated by one of the rationale providers is better than the other and if yes, select the better one. Based on the answer to Question 2, we also estimate if the better rationale adjudged by human evaluator aligns with likelihood-based rationale selection. We found that for 1) 72\% cases, \approachName{}'s rationales are meaningful and helpful; 2) for 78.67\% cases, one rationale is better than the other indicating that rationales from the two rationale providers are distinct and diverse; and 3) in 84\% cases, better rationale judged by humans aligns with likelihood-based rationale selection.

\noindent \textbf{Qualitative Error Analysis:} We discuss the errors made by \approachName{} to understand the scope of further improvements. Consider the following \textbf{sample} from GSM8K - \textit{``Elsa started the day with 40 marbles. At breakfast, she lost 3 marbles. At lunchtime, she gave her best friend Susie 5 marbles. In afternoon, Elsa's mom bought her a new bag with 12 marbles. Susie came back and gave Elsa twice as many marbles as she received at lunch. How many marbles did Elsa end the day with?"}. The \textbf{correct logic} would be - \textit{``Elsa had 40 (marbles) - 3 - 5 = 32 marbles. Elsa then gained 12 marbles + 32 = 44 marbles. At dinner Elsa received 2 * 5 = 10 marbles from Susie. In total, Elsa had 44 marbles + 10 = 54"}. The rationale generated by the \textbf{first rationale provider ($\mathcal{RP}_1$)} is - \textit{``Elsa started the day with 40 - 3 = 37 marbles. She gave Susie \textcolor{red}{5 + 12 = 17 marbles}. Susie then gave Elsa \textcolor{red}{17 * 2 = 34 marbles}. In total, Elsa ended the day with \textcolor{red}{37 + 17 + 34 = 88 marbles}"}. The rationale generated by the \textbf{second rationale provider ($\mathcal{RP}_2$)} is - \textit{``Elsa started the day with 40 marbles. At breakfast, she lost 3 marbles while playing \textcolor{darkgreen}{(40 - 3)}. At lunchtime, she gave her best friend Susie 5 marbles (40 - 3 \textcolor{red}{+} 5). In the afternoon, Elsa's mom bought her a new bag with 12 marbles (40 - 3 \textcolor{red}{+} 5 \textcolor{darkgreen}{+} 12). Susie came back and gave Elsa twice as many marbles as she received at lunch (40 - 3 \textcolor{red}{+} 5 \textcolor{darkgreen}{+ 12 + 2*5})"}. This example demonstrates that the rationale generated by $\mathcal{RP}_2$ is almost correct except that while incorporating the fact that Elsa gave 5 marbles, it uses a ‘+’ sign instead of ‘-’. This indicates that the rationale from a rationale provider might only be slightly incorrect and minor refinement can make it correct. Thus, rationale refinement can be explored in future work. Further, employing more rationale providers can increase the chances of generating the correct rationale which can be used for DPO.


\noindent \textbf{Additional Experiments:} We treat number of rationale providers (RPs) as a hyper-parameter and show that employing 3 RPs improve accuracy further (Appendix~\ref{sec:app_more_rps}). We observe that rationales from \approachName\ are effective through perplexity estimation (Appendix~\ref{app:add_eval_rationales}). Appendix~\ref{app:comp_complexity} discusses that \approachName's computation load is shifted to train time and inference requires generating a single rationale.

\subsection{Impact of Design Choices for \approachName}
\label{app:abl_studies}
\begin{table*}[t]
    \centering
    \resizebox{\textwidth}{!}{%
    \begin{tabular}{ccccccccc}
        \toprule
        \multirow{2}{*}{\textbf{Ablation ID}} & \textbf{Distinct Rationale} & \textbf{Likelihood-based} & \textbf{DPO Sample} & \multicolumn{1}{c}{\textbf{Maths}}  & \multicolumn{2}{c}{\textbf{NLI}} & \multicolumn{2}{c}{\textbf{Comonsense}} \\
        & \textbf{Providers} & \textbf{Rationale Selection} & \textbf{Filtration} & \textbf{GSM8K} & \textbf{WinoGrande} & \textbf{PIQA} & \textbf{HellaSwag} & \textbf{CSQA}  \\ \midrule
         1 & No & Yes & Yes & 38.22 & 40.61 & 45.17 & 94.15 & 72.45 \\
         2 & Yes & No & Yes & 47.16 & 44.62 & 53.17 & 95.27 & 76.39\\
         3 & Yes & Yes & No & 42.11 & 42.75 & 48.25 & 94.36 & 73.18\\
         4 & Yes & No & No & 40.26 & 41.73 & 48.52 & 92.11 & 72.13 \\
         5 & No & No & No & 39.13 & 41.44 & 47.41 & 93.26 & 71.77\\ \midrule
         \textbf{\approachName} & Yes & Yes & Yes & \textbf{53.48} & \textbf{53.49} & \textbf{59.57} & \textbf{99.21} & \textbf{80.13} \\
        \bottomrule
    \end{tabular}%
    }
    \caption{Ablation study showing importance of \approachName's components - \textbf{A) Distinct Rationale Providers:} Sampling-based decoding is used for diverse outputs as an alternative; \textbf{B) Likelihood-based rationale selection}: LLM-as-a-judge is used to rate rationales; \textbf{C) Sample filtration}: Entire train set is used for DPO w/o filtration.}
    \label{tab:ablation}
\end{table*}
\begin{table*}[t]
\centering
\resizebox{\textwidth}{!}{%
\begin{tabular}{lccccc}
\toprule
\multirow{2}{*}{\textbf{Model}} & \multicolumn{1}{c}{\textbf{Maths}}  & \multicolumn{2}{c}{\textbf{NLI}} & \multicolumn{2}{c}{\textbf{Comonsense}} \\
        & \textbf{GSM8K} & \textbf{WinoGrande} & \textbf{PIQA} & \textbf{HellaSwag} & \textbf{CSQA}  \\
\midrule
GPT-4o (3-shot)                & 77.84 & 88.47 & 83.38 & 95.49 & 92.97 \\
\midrule
LLaMA3-8B~\citep{dubey2024llama} & 60.68 (-17.16) & 48.21 (-40.26) & 49.35 (-34.03) & 90.35 (-5.14) & 80.39 (-12.58) \\
        \hspace{5mm} \textbf{w/ \approachName\ (ours)} & \textbf{69.09} (-8.75) & \textbf{54.41} (-34.06) & \textbf{58.70} (-24.68) & \textbf{92.86} (-2.63) & \textbf{87.42} (-5.55) \\
\bottomrule
\end{tabular}%
}
\caption{Comparison of 3-shot evaluation of GPT-4o with LLaMA3-8B with and without \approachName. We see a significant reduction in the performance gap between GPT-4o and LLaMa3-8B when used with \approachName.}
\label{tab:comp_perf_gap}
\end{table*}
We examine the effectiveness of following components (using open-llama-v2-7b backbone) - 1) Distinct Rationale Providers, 2) Task-guided Likelihood-based Rationale Selection, and 3) Sample Filtration for DPO. Table~\ref{tab:ablation} shows the results where it can be seen that obtaining diverse rationales from distinct rationale providers gives significantly better performance than sampling-based decoding using a single model \textbf{(\approachName\ vs. row 1)}. To analyse the importance of using likelihood of generating the GT answer to rate a rationale, we leverage LLM-as-a-judge paradigm (using IFT version of backbone LLM) as an alternative where the LLM is prompted to rate rationales \textbf{(row 2)}. Notable drop in performance than \approachName\ indicates that prompt-based rating does not work at scale of small LMs. Likewise, filtration of samples for task-guided DPO conditioned on whether best rationale enhances likelihood of generating GT (than not using any rationale) is critical for gains achieved by \approachName\ \textbf{(vs. row 3)}. Omitting both likelihood-based rationale selection and sample filtration \textbf{(row 4)} as well as all the three components \textbf{(row 5)} gives significantly lower performance.

\subsection{Bridging Accuracy Gap between Smaller Models and GPT-4o using COLLATE}
We evaluate the performance of GPT-4o for different tasks (using 3-shot prompting). We then compare the accuracy gap of Llama3-8B from GPT-4o and report reduction in this gap using rationales from COLLATE (with Llama3-8B) in Table~\ref{tab:comp_perf_gap}. For GSM8K, the performance gap of Llama3-8B with Gpt-4o reduces from 17.16\% to 8.75\% which is a significant relative reduction of 49\%. Likewise, it reduces from 40.26\% to 34.06\% (relative reduction of 15\%) on WinoGrande, 34\% to 24\% (relative reduction of 27.48\%) on PIQA, 5\% to 2.63\% on HellaSwag, and from 12.58\% to 5.55\% on CSQA.


\section{Conclusion}
\label{sec:conclusion}
Step-by-step rationales generated by large LMs have been commonly used to improve smaller LMs. Such large-scale (often closed) LLMs cannot be used to train other models in commercial setting owing to legal constraints. Little focus has been given to enhancing rationale generation ability of small LMs. We propose \approachName\ which optimises an LLM to selectively generate better rationales from a pool of diverse candidates produced by distinct instances of the same LLM. Rationale candidates are ranked for preference tuning based on end-task-guided likelihood score. \approachName\ outperforms multiple baselines on diverse tasks without relying on larger LMs to generate and rate rationales. It provides significant accuracy gains for LMs of different parameter-scales and families. 


\section{Limitations}
To obtain the distinct rationale providers, currently we randomly divide the training set into as many equal splits as the number of distinct rationale providers. However, it could be explored if the samples in the train dataset could be allocated to each rationale provider in a way such that each rationale provider gains expertise in certain task domain(s) compared to the other rationale providers. This way, the different rationale providers would automatically become specialized experts for different types of problems and tasks. This can be used to obtain insights into which type of experts are required to improve rationales for different task domains. Additionally, the number of distinct rationale providers to be used a hyper-parameter. We experiment with two rationale providers in the main paper and also show that employing 3 rationale providers improve the accuracy gains further (in Appendix~\ref{sec:app_more_rps}, Table~\ref{tab:collate_rps_comparison}). However, it could be explored if the optimal number of rationale providers could be determined automatically for a given task.

\bibliography{acl_latex}

\appendix

\section{Qualitative Analysis and Comparison of Rationales}
\label{sec:app_qualitative}

We discussed the qualitative comparison of rationales generated by \approachName\ with the best baseline in the section - `Comparison of COLLATE with Baselines' (Table 2) in the main paper for one dataset from each task domain. Table~\ref{tab:qualitative_samples} shows more examples comparing the quality of rationales across all the datasets for different task domains.

\begin{table*}[t]
    \centering
    \resizebox{\textwidth}{!}{%
    \begin{tabular}{c|p{200mm}}
        \hline
        \textbf{Dataset} & \textbf{Task-Instruction and Rationales Generated by Best Baseline (SPIN) and Rationale Providers - $\mathcal{RP}_1$ and $\mathcal{RP}_2$ in \approachName} \\ \hline
        \multirow{4}{*}{GSM8K} & \textbf{Instruction:} I have 10 liters of orange drink that are two-thirds water and I wish to add it to 15 liters of pineapple drink that is three-fifths water. But as I pour it, I spill one liter of the orange drink. How much water is in the remaining 24 liters?\\ 
        & \textbf{Baseline:} The orange drink is two-thirds water, so 10 liters would have 6.5 liters of water. The pineapple drink is three-fifths water, so 15 liters would have 9 liters of water. Since 1 liter of the orange drink is spilled, only 5.5 liters of water from the orange drink remains. Adding it to the 9 liters from the pineapple drink gives a total of 14.5 liters of water in the remaining 24 liters. \textcolor{red}{\xmark}\\
        & \textbf{Rationale Provider 1:} The orange drink is two-thirds water. When 1 liter is spilled, the remaining 9 liters of orange drink contains 2/3x9=6 liters of water. The pineapple drink is three-fifths water, so 15 liters contains 9 liters of water. The total water content in 24 liters is 6+9=15 liters of water. \textcolor{teal}{\cmark}\\ 
        & \textbf{Rationale Provider 2:} Since the orange drink is two-thirds water, spilling 1 liter reduces the water content by 1 liter, leaving 2/3 x 9 liters of water. The pineapple drink has 9 liters of water. Therefore, the total amount of water should be 9+9=18 liters of water in the remaining 24 liters. \textcolor{red}{\xmark}\\
        \hline
        \multirow{4}{*}{WinoGrande} & \textbf{Instruction:} Terry tried to bake the eggplant in the toaster oven but the \_\_\_\_\_\_\_\_ was too big. A. eggplant, B. toaster\\
        & \textbf{Baseline:} The eggplant didn't fit properly in the toaster. Therefore, the correct answer seems to be ``toaster". \textcolor{red}{\xmark}\\
        & \textbf{Rationale Provider 1:} The issue described is that the eggplant didn't fit in the toaster. Therefore, the correct answer is ``eggplant" because it was too big for the toaster.\textcolor{teal}{\cmark}\\
        & \textbf{Rationale Provider 2:} Since the toaster is the appliance being used, it might have been too large to handle for baking the eggplant. So, the correct answer could be ``toaster". \textcolor{red}{\xmark}\\
        \hline
        \multirow{4}{*}{PIQA} & \textbf{Instruction:} To make a bedroom temperature colder without a fan or air conditioner, sol1: hang up a thick, and dark blanket over the windows and then ensure their is proper air flow in the room. sol2: wave a large blanket up and down to generate a gust of wind to cool the room down when it gets too hot.\\
        & \textbf{Baseline:} Waving a large blanket up and down would create a gust of wind, which might help cool the room down quickly. Therefore, sol 2 is correct. \textcolor{red}{\xmark}\\ 
        & \textbf{Rationale Provider 1:} Generating wind by waving a blanket can circulate air and help cool the room, especially when it gets too hot. Therefore, sol 2 is likely the more effective method for cooling the room. \textcolor{red}{\xmark} \\
        & \textbf{Rationale Provider 2:} To cool down a room, blocking out sunlight and reducing heat entry is essential. Hanging a thick, dark blanket over the windows prevents heat from entering, which helps keep the room cooler. Therefore, sol 1 the correct answer. \textcolor{teal}{\cmark}\\
        \hline
        \multirow{4}{*}{HellaSwag} & \textbf{Instruction:} Then he takes a small stone from the flowing river and smashes it on another stone. He starts to crush the small stone to smaller pieces. He \_\_\_\_\_\_\_\_\_\_\_\_\_\_\_\_. A. cuts the center stone in half and blow it on to make it bigger. B. grind it hard to make the pieces smaller, C. eventually brings it back into view and adds it to the smaller ones to make a small triangular shaped piece, D. starts to party with them and throw the pieces by hand while they celebrate.\\
        & \textbf{Baseline:} \textcolor{red}{\xmark} Once the small stone is crushed, we need to start to party with them and throw the pieces by hand while they celebrate. This is an unexpected, yet possible action, making option D correct. \textcolor{red}{\xmark}\\
        & \textbf{Rationale Provider 1:} After crushing the stone into smaller pieces, one should grind it hard to make the pieces smaller. So, option B is the correct answer. \textcolor{teal}{\cmark}\\
        & \textbf{Rationale Provider 2:} After crushing the stone into smaller pieces, it makes sense to cut the center stone in half and blow it on to make it bigger. So option A seems correct. \textcolor{red}{\xmark}\\
        \hline 
        \multirow{4}{*}{CSQA} & \textbf{Instruction:} If I was getting drunk, and people couldn't understand me, what might I be having? A. a seizure, B. slurred speech, C. death, D. forgetfulness, E. pass out\\
        & \textbf{Baseline:} Getting drunk could lead to serious consequences like ``death" if things get out of hand. If people can't understand you, it could be a sign that something is seriously wrong, making option C correct. \textcolor{red}{\xmark} \\
        & \textbf{Rationale Provider 1:} EWhen you're getting drunk, the most common reason people can't understand you is ``slurred speech." Alcohol often impairs speech clarity, making option B the correct answer. \textcolor{teal}{\cmark}\\
        & \textbf{Rationale Provider 2:} If people can't understand you while you're getting drunk, it might be because you're experiencing ``forgetfulness." Alcohol can affect memory, so you might be forgetting words or how to speak clearly. Therefore, option D seems correct. \textcolor{red}{\xmark}\\
        \hline
    \end{tabular}%
    }
    \caption{Qualitative comparison of rationales generated by best baseline (SPIN) and rationale providers in \approachName\ on all datasets from each task domain. \approachName\ generates better rationales by virtue of employing distinct rationale providers.}
    \label{tab:qualitative_samples}
\end{table*}

\begin{table*}[t]
    \centering
    \resizebox{\textwidth}{!}{%
    \begin{tabular}{c|p{200mm}}
        \hline
        \textbf{Dataset} & \textbf{Task-Instruction and Ground-Truth Answers for Each Dataset used to evaluate \approachName} \\ \hline
        \multirow{2}{*}{GSM8K} & \textbf{Instruction:} Betty is saving money for a new wallet which costs \$100. Betty has only half of the money she needs. Her parents decided to give her \$15 for that purpose, and her grandparents twice as much as her parents. How much more money does Betty need to buy the wallet?\\ 
        & \textbf{Ground-Truth Response:} 5\\
        \hline
        \multirow{2}{*}{WinoGrande} & \textbf{Instruction:} Terry tried to bake the eggplant in the toaster oven but the \_\_\_\_\_\_\_\_ was too big. A. eggplant, B. toaster\\
        & \textbf{Ground-Truth Response: } A. eggplant \\
        \hline 
        \multirow{2}{*}{PIQA} & \textbf{Instruction:} How to dry flowers? sol1: Find a dark, moist area with good circulation, such as an attic or unused closet. With unflavored dental floss, secure the bottom of the flowers' stems to a hanger so that they hang upside down to dry. Leave flowers for two to three weeks until completely dry, sol2: Find a dark, dry area with good circulation, such as an attic or unused closet. With unflavored dental floss, secure the bottom of the flowers' stems to a hanger so that they hang upside down to dry. Leave flowers for two to three weeks until completely dry.\\
        & \textbf{Ground-Truth Response: } sol2\\
        \hline 
        \multirow{2}{*}{HellaSwag} & \textbf{Instruction:} Then he takes a small stone from the flowing river and smashes it on another stone. He starts to crush the small stone to smaller pieces. He \_\_\_\_\_\_\_\_\_\_\_\_\_\_\_\_. A. cuts the center stone in half and blow it on to make it bigger. B. grind it hard to make the pieces smaller, C. eventually brings it back into view and adds it to the smaller ones to make a small triangular shaped piece, D. starts to party with them and throw the pieces by hand while they celebrate.\\
        & \textbf{Ground-Truth Response: } B\\
        \hline 
        \multirow{4}{*}{CSQA} & \textbf{Instruction:} When learning about the world and different cultures, what is important if you are committed to eliminating preconceived notions. A. newness, B. loss of innocence, C. enlightenment, D. open mind, E. smartness\\
        & \textbf{Ground-Truth Response: } D. open mind\\
        \hline
    \end{tabular}%
    }
    \caption{Examples of instructions from different datasets belonging to diverse task domains used in the experiments - (i) Maths Problem Solving (GSM8K), (ii) Natural Language Inference (WinoGrande and PIQA), and (iii) Commonsense Reasoning (HellaSwag and CSQA).}
    \label{tab:dataset_samples}
\end{table*}

\section{Prompt Templates for Multi-Mode Instruction Fine-Tuning}
\label{sec:app_prompt_template}
As discussed in the Methodology section in the main paper, the base model $\mathcal{M}$ is instruction fine-tuned to enable the LLM to operate in three modes - (i) generate the final answer given the instruction as input ($\mathcal{I} \rightarrow \mathcal{A}$); (ii) generate the rationale given the instruction as input ($\mathcal{I} \rightarrow \mathcal{R}$); and (iii) generate the answer conditioned on the instruction and rationale as input ($[\mathcal{I} ; \mathcal{R}] \rightarrow \mathcal{A}$). The inputs to the LLM are formatted using corresponding prompts ($\mathcal{P}_{\mathcal{I} \rightarrow \mathcal{A}}$; $\mathcal{P}_{\mathcal{I} \rightarrow \mathcal{R}}$; $\mathcal{P}_{[\mathcal{I} ; \mathcal{R}] \rightarrow \mathcal{A}}$) for each of these modes so that the LLM can generate an appropriate output accordingly. The textual instruction for each prompt template is specified as follows:

\begin{enumerate}
    \item $\mathcal{P}_{\mathcal{I} \rightarrow \mathcal{A}} = $ ``You are an AI assistant `M'. Provide a response to the given instruction denoted by Task Description.\\ \\ 
    \lbrack TASK DESCRIPTION STARTS\rbrack\\ 
    \textlangle Task Description\textrangle: In this task, you will be given an `Instruction'. Generate the correct answer for the given instruction.\\
    `Instruction' - \textlangle instruction\textrangle\\
    \lbrack TASK DESCRIPTION ENDS\rbrack\\ \\
    For the given \textlangle Task Description\textrangle, give your response. [M RESPONSE BEGINS]: "
    
    \item $\mathcal{P}_{\mathcal{I} \rightarrow \mathcal{R}} = $ ``You are an AI assistant `M'. Provide a response to the given instruction denoted by Task Description.\\ \\ 
    \lbrack TASK DESCRIPTION STARTS\rbrack\\ 
    \textlangle Task Description\textrangle: In this task, you will be given an `Instruction'. Generate descriptive reasoning on how to derive the correct answer for the instruction such that the descriptive reasoning will be useful to another AI assistant to generate the correct answer.\\
    `Instruction' - \textlangle instruction\textrangle\\
    \lbrack TASK DESCRIPTION ENDS\rbrack\\ \\
    For the given \textlangle Task Description\textrangle, give your response. [M RESPONSE BEGINS]: "

    \item $\mathcal{P}_{[\mathcal{I} ; \mathcal{R}] \rightarrow \mathcal{A}} = $ ``You are an AI assistant `M'. Provide a response to the given instruction denoted by Task Description.\\ \\ 
    \lbrack TASK DESCRIPTION STARTS\rbrack\\ 
    \textlangle Task Description\textrangle: In this task, you will be given an `Instruction' and a rationale denoted by `Rationale'. Analyse the rationale and come up with the correct answer for the given instruction.\\
    `Instruction' - \textlangle instruction\textrangle\\
    `Rationale' - \textlangle rationale\textrangle\\
    \lbrack TASK DESCRIPTION ENDS\rbrack\\ \\
    For the given \textlangle Task Description\textrangle, give your response. [M RESPONSE BEGINS]: "
\end{enumerate}

\noindent In the above prompt templates, \textlangle instruction\textrangle\ is a placeholder for the actual task instruction $\mathcal{I}^{\mathcal{T}}$ and \textlangle rationale\textrangle\ is a placeholder for the rationale text.


\section{Direct Preference Optimisation (DPO)}
\label{sec:app_dpo}
Direct Preference Optimisation (DPO)~\citep{NEURIPS2023_a85b405e} was introduced as an alternative to Reinforcement Learning using Human Feedback (RLHF)~\citep{NEURIPS2022_b1efde53} technique to alleviate the need of training a reward model. RLHF depends on training a reward model to assign a score to the outputs generated by an LLM to fine-tune the LLM through reinforcement learning to align it with human preferences. On the other hand, DPO transforms the loss over the reward-function to a loss over the LLM policy such that the reward is optimised implicitly by optimising the loss over the policy. It does so by leveraging human preference data which compares two possible outputs generated by an LLM such that the better output is considered as the winner candidate - $y_w$ while the inferior output is considered as the loser candidate - $y_l$. Given a static dataset of the form $\mathcal{D} = \{x, y_w, y_l\}$, where x is the input, the loss is modeled as - 

\begin{align}
    \mathcal{L}_{R} = -log[\sigma(r(x, y_w) - r(x, y_l))] \\
    r(x, y) = \beta log(\frac{\pi_{\theta}(y|x)}{\pi_{ref}(y|x)})
\end{align}

where, $\pi_{\mathcal{Z}}(y | x)$ is the likelihood of generating $y$ given $x$ as input to the model $\mathcal{Z} \in \{\mathcal{M}_{ref}, \mathcal{M}_{\theta}\}$, $\mathcal{M}_{ref}$ is usually taken to be the instruction fine-tuned model in the case of an LLM to prevent the LLM policy from deviating too much from the initial policy, $\mathcal{M}_{\theta}$ represents the LLM policy being optimised through DPO, $\sigma$ is the sigmoid activation, and $\beta$ is a coefficient that controls the amount of deviation from the reference model. In summary, the algorithm optimises the LLM to learn to prefer generating certain outputs over other candidates without requiring an explicit reward model. Please refer to the original publication~\citep{NEURIPS2023_a85b405e} for an elaborate discussion of the details.

\begin{table*}[t]
\centering
\begin{tabular}{lccccc}
\toprule
\multirow{2}{*}{\textbf{Model}} & \multicolumn{1}{c}{\textbf{Maths}}  & \multicolumn{2}{c}{\textbf{NLI}} & \multicolumn{2}{c}{\textbf{Comonsense}} \\
        & \textbf{GSM8K} & \textbf{WinoGrande} & \textbf{PIQA} & \textbf{HellaSwag} & \textbf{CSQA}  \\
\midrule
Llama-3-8B-Base            & 52.16 & 79.43 & 73.16 & 69.21 & 60.07  \\
Llama-3-8B-Base w COLLATE  & 54.49 & 81.37 & 77.02 & 72.16 & 63.15  \\
\midrule
Llama-3-8B-Instruct        & 76.12 & 78.51 & 72.06 & 76.25 & 57.69  \\
Llama-3-8B-Instruct w COLLATE & 81.26 & 81.22 & 75.69 & 80.19 & 59.14  \\
\bottomrule
\end{tabular}
\caption{Few-shot evaluation of LLaMa-3-8B models (Base and Instruct) on Maths (GSM8K), NLI (WinoGrande, PIQA), and Commonsense reasoning (HellaSwag, CSQA). Results show that \approachName\ outperforms corresponding baseline, with notable gains in both NLI and Commonsense tasks.}
\label{tab:llama_results}
\end{table*}

\begin{table*}[t]
    \centering
    \resizebox{0.7\textwidth}{!}{%
    \begin{tabular}{cccccc}
        \toprule
        \multirow{2}{*}{\textbf{DPO Iteration}} & \multicolumn{1}{c}{\textbf{Maths}}  & \multicolumn{2}{c}{\textbf{NLI}} & \multicolumn{2}{c}{\textbf{Comonsense}} \\
        & \textbf{GSM8K} & \textbf{WinoGrande} & \textbf{PIQA} & \textbf{HellaSwag} & \textbf{CSQA}  \\ \midrule
        Iteration 0 & 47.95 & 50.86 & 49.84 & 98.93 & 78.74\\
        Iteration 1 & 53.41 & 51.65 & 55.33 & 99.21 & 80.11\\
        Iteration 2 & \textbf{53.48} & \textbf{53.49} & \textbf{59.57} & \textbf{99.21} & \textbf{80.13}\\
        \bottomrule
    \end{tabular}%
    }
    \caption{Performance analysis of rationale provider (selected for task-specific SFT-based evaluation) across different iterations of DPO training using open-llama-v2-7b backbone. It can be seen that performance increases with iterations for all the tasks.}
    \label{tab:across_dpo_iterations}
\end{table*}

\section{Dataset Samples}
\label{sec:app_dataset_samples}
Details of datasets were discussed in the `Experiments and Evaluation' section in the main paper. Table~\ref{tab:dataset_samples} in the appendix shows samples of instructions for each dataset from all task domains - (i) Maths Problem Solving (GSM8K), (ii) Natural Language Inference (WinoGrande and PIQA), and (iii) Commonsense Reasoning (HellaSwag and CSQA).

\section{Few-shot evaluation of \approachName}
\label{app:few_shot_eval}
The few-shot evaluation results in Table \ref{tab:llama_results} demonstrate that COLLATE significantly improves performance across tasks. On GSM8K, the base model rose from 52.16 to 54.49 (+2.33), and the instruct model from 76.12 to 81.26 (+5.14). For NLI, COLLATE boosted WinoGrande by +1.94 (base) and +2.71 (instruct), and PIQA by +3.86 (base) and +3.63 (instruct). In commonsense reasoning, HellaSwag improved by +2.95 (base) and +3.94 (instruct), while CSQA saw smaller gains of +3.08 (base) and +1.45 (instruct). These results highlight COLLATE's consistent effectiveness in enhancing few-shot performance across diverse benchmarks.

\section{Performance of Rationale Providers across Iterations}
\label{app:across_iterations}

Table~\ref{tab:across_dpo_iterations} shows the performance analysis of rationale provider (selected for task-specific SFT-based evaluation) across different iterations of DPO training using open-llama-v2-7b backbone. It can be seen that performance increases with iterations for all the tasks.

\section{More Rationale Providers (RPs) improve accuracy further}
\label{sec:app_more_rps}
Table \ref{tab:collate_rps_comparison} shows that using 3 RPs improves performance over the configuration with 2 RPs across GSM8K and PIQA. Specifically, on GSM8K, COLLATE with 2 RPs achieves an accuracy of 53.58, while using 3 RPs boosts this to 55.24, a gain of 1.66\%. Similarly, for PIQA, the accuracy improves from 59.57 (with 2 RPs) to 61.73 (with 3 RPs), reflecting a 2.16\% increase. These results demonstrate that increasing the number of RPs further enhances the accuracy, showing that COLLATE benefits from the diversity and richness of additional rationales. This improvement over baselines is further amplified when using 3 RPs, highlighting the efficacy of \approachName.

\begin{table}[h]
\centering
\begin{tabular}{lcc}
\toprule
\textbf{Method} & \textbf{GSM8K} & \textbf{PIQA} \\
\midrule
COLLATE w 2 RPs  & 53.58 & 59.57 \\
COLLATE w 3 RPs  & 55.24 & 61.73 \\
\bottomrule
\end{tabular}
\caption{Impact of Using Multiple Rationale Providers (RPs) in COLLATE. This table compares the performance of COLLATE with 2 and 3 rationale providers (RPs) on the GSM8K and PIQA benchmarks, demonstrating that using 3 RPs leads to further improvements in accuracy for the open-llama-v2-7B backbone.}
\label{tab:collate_rps_comparison}
\end{table}

\begin{table*}[t]
\centering
\begin{tabular}{lccccc}
\toprule
\multirow{2}{*}{\textbf{Model}} & \multicolumn{1}{c}{\textbf{Maths}}  & \multicolumn{2}{c}{\textbf{NLI}} & \multicolumn{2}{c}{\textbf{Comonsense}} \\
        & \textbf{GSM8K} & \textbf{WinoGrande} & \textbf{PIQA} & \textbf{HellaSwag} & \textbf{CSQA}  \\
\midrule
CoT                       & 35.1 & 18.5 & 17.7 & 28.4 & 22.9 \\
EoT                       & 121.5 & 66.1 & 69.4 & 75.9 & 84.1 \\
Distilling Step-by-Step   & 36.1 & 19.0 & 18.4 & 28.1 & 23.1 \\
SPIN                      & 43.2 & 22.7 & 22.2 & 33.3 & 27.5 \\
\midrule
\textbf{COLLATE}                   & 48.1 & 25.7 & 24.9 & 37.1 & 30.8 \\
\bottomrule
\end{tabular}
\caption{Comparison of Average Number of Rationale Tokens Generated During Inference. \approachName\ demonstrates no significant increase in inference-time overhead compared to baselines, as its computational cost is primarily shifted to the training phase where diverse rationales are generated using multiple rationale providers ($RP$s) for DPO. In contrast, prompt-based baselines incur higher inference costs as they require generating multiple rationales during inference.}
\label{tab:comp_complexity}
\end{table*}
\begin{table*}[t]
\centering
\begin{tabular}{lccccc}
\toprule
\multirow{2}{*}{\textbf{Model}} & \multicolumn{1}{c}{\textbf{Maths}}  & \multicolumn{2}{c}{\textbf{NLI}} & \multicolumn{2}{c}{\textbf{Comonsense}} \\
        & \textbf{GSM8K} & \textbf{WinoGrande} & \textbf{PIQA} & \textbf{HellaSwag} & \textbf{CSQA}  \\
\midrule
SFT w/o rationales             & 12.93 & 3.39 & 5.86 & 8.4  & 9.77 \\
SFT w EoT rationales           & 12.61 & 3.28 & 5.59 & 7.92 & 9.04 \\
SFT w Distilling Step-by-Step  & 11.47 & 3.15 & 5.36 & 7.33 & 8.48 \\
SFT w SPIN rationales          & 10.28 & 3.10 & 5.27 & 7.17 & 8.09 \\
\midrule
SFT w COLLATE rationales       & 8.03  & 2.53 & 4.79 & 6.26 & 7.72 \\
\bottomrule
\end{tabular}
\caption{Perplexity Comparison of \approachName\ and Baseline Methods for generating ground-truth (GT) answers conditioned on rationales in the input, as evaluated by task-specific SFT models (Llama-3-8B). COLLATE achieves the lowest perplexity across all tasks, demonstrating its effectiveness in providing rationales that enhance the model's certainty in generating correct answers.}
\label{tab:perplexity_comparison}
\end{table*}

\section{Computational Complexity - Average number of Rationale tokens}
\label{app:comp_complexity}
Table \ref{tab:comp_complexity} highlights the average number of rationale tokens generated during inference for COLLATE and baseline methods. COLLATE maintains comparable inference-time efficiency, with no significant increase in rationale tokens compared to baselines. For example, COLLATE generates an average of 33.3 tokens per rationale, similar to baselines like CoT (24.5 tokens) and SPIN (29.8 tokens). EoT, being a prompt-based method employs multiple LLMs during inference, thereby shooting the average number of rationale tokens to 83.4

The efficiency of \approachName\ is due to its design, which shifts the computational overhead to the training phase—a one-time process. During training, multiple rationale providers ($RP$s) are used to generate diverse rationales for DPO, ensuring high-quality rationale generation. However, at inference time, \approachName\ employs only a single $RP$ to generate a single rationale, minimizing additional computation. This contrasts with prompt-based baselines, which require generating multiple rationales during inference, leading to higher computational costs. By reducing inference-time overhead, \approachName\ is particularly well-suited for practical applications where computational resources and latency are critical. This efficiency, coupled with its robust performance, makes \approachName\ a better alternative for deploying models in real-world settings.

\section{Additional Evaluation of Rationales from \approachName}
\label{app:add_eval_rationales}
As an additional evaluation metric, we estimate the perplexity of generating ground-truth (GT) answers conditioned on rationales in the input for test samples. This assessment was conducted using task-specific SFT models (open-llama-v2-7B), and Table \ref{tab:perplexity_comparison} compares COLLATE against the best-performing baseline from each category.

The results indicate that COLLATE achieves the lowest perplexity across all tasks, highlighting its ability to produce rationales that better guide the model towards correct answers. For instance, on GSM8K, COLLATE achieves a perplexity of 8.03, compared to 10.28 for the best baseline (SPIN). Similarly, on CSQA and HellaSwag, COLLATE records perplexities of 7.72 and 6.26 respectively, outperforming SPIN with scores of 8.09 and 7.17. This improvement underscores the effectiveness of COLLATE rationales in reducing model uncertainty, making it superior to other methods. The results reaffirm COLLATE’s ability to provide rationales that not only enhance model performance but also ensure higher confidence in the answers, as evidenced by consistently lower perplexity scores.

\section{Selection of Rationale Provider for Evaluation}
\label{app:rp_selection}

\begin{figure}[h] 
    \centering 
    \includegraphics[width=\columnwidth]{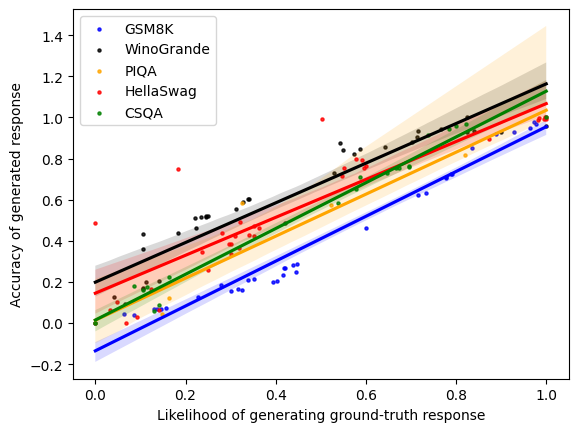} 
    \caption{Normalized plot between task test accuracy vs. likelihood of generating GT on CoT val-split. Positive correlation indicates that rationale provider selected based on better likelihood is suitable choice to generate rationales for task-specific SFT-based evaluation.}
    \label{fig:corr_like_acc} 
\end{figure}

We use a single rationale provider while assessing rationales generated by \approachName. The best rationale provider is selected based on likelihood estimation on CoT val-split samples. We measure probability of generating ground-truth answer by the IFT model $\mathcal{M}_{IFT}$ conditioned on the rationale. The rationale provider that yields maximum value (averaged across samples) is used for evaluation. Figure~\ref{fig:corr_like_acc} shows correlation between test-accuracy and likelihood indicating that rationale provider with best likelihood is a suitable choice.

\section{Human Study for Rationale Evaluation}
\label{app:human_study_app}
We conducted a human study to evaluate the effectiveness of rationales obtained using the proposed \approachName{} framework. The following steps describe creation of data for human evaluation:

\textbf{Dataset Creation for Human Evaluation}

\begin{enumerate}
    \item We collected a total of 75 samples based on random selection by taking an equal number of samples for each task i.e. 15 samples randomly from the test sets of each of the 5 task datasets.
    \item For each sample, we obtain the rationales $R1_g$, $R2_g$ from the two LLM rationale providers.
\end{enumerate}

Once the above rationales are obtained, we employed three paid human annotators and presented them with the instruction in each sample along with different rationales obtained above. The human evaluators are asked to judge the quality of different rationales based on the following questions and guidelines:

\textbf{Questions and Guidelines}

\begin{enumerate}
    \item Question 1: Is the final rationale obtained from COLLATE useful for answering the question correctly? The rationale is useful if it is correct and provides the correct explanation on how the answer for the instruction in the sample should be derived. Provide a label out of 0 or 1 such that 0 means that the final rationale is totally wrong; and 1 means that the final rationale is totally correct.
    \item Question 2: Compare the two rationales obtained using the two rationale providers - $R1_g$ and $R2_g$. Provide a label of 0 or 1 where 0 means that none of the rationales is better than the other and 1 means that one rationale is better than the other.
    \item Question 3: In Question 2, in case one rationale is better than the other (between the rationales obtained from the two rationale providers), select the better rationale. 
\end{enumerate}

\begin{table*}[t]
    \centering
\begin{tabular}{lc}
\hline
Metric Name & Value (in \%) \\ 
\hline
 Final Rationale Alignment  &  72.0  \\ 
 Diversity b/w two Rationales  &  78.67  \\ 
 Better Rationale Alignment with Likelihood-based Selection  &  84.0  \\ 
\hline
\end{tabular}
\caption{Human study results summarizing values of different metrics evaluated using human labels. It is observed that for good proportion of cases, final rationale obtained from \approachName\ aligns with human preferences, the rationales obtained from two variants are diverse and better rationale judged by humans matches with winner rationale selected using likelihood based utility score.}
\label{tab:app_human_eval_results}
\end{table*}

\textbf{Definition of Metrics Estimated from Human Labels}

Different rationales were presented to human evaluators in jumbled order to avoid biases while comparing rationales. Based on the judgement labels provided by the human evaluators for 3 questions above for the 75 samples, we estimate the following metrics: 

\begin{enumerate}
    \item Final Rationale Alignment – \% proportion of samples which were assigned label 1 i.e. totally correct. 
    \item Diversity b/w two Rationales - \% proportion of samples where the two rationales $R1_g$ and $R2_g$ obtained from two rationale providers are different i.e. cases where one of the two rationales is better than the other (label 1). This metric is estimated to verify if the variants truly generate distinct rationales.
    \item Better Rationale Alignment with Likelihood based Selection: We consider samples where label 1 is provided to Question 2 i.e. one of the generated rationales is judged better than the other generated rationale (comparing $R1_g$ and $R2_g$). We estimate the metric as \% proportion cases from these samples where better rationale determined using likelihood-based utility score matches the better rationale from human judgement.
\end{enumerate}

\textbf{Human Study Results and Discussion}

We compute the above metrics using the 75 samples used for human evaluation. We report the average of metrics obtained for the three human evaluators in Table~\ref{tab:app_human_eval_results}. We discuss following observations from the results in Table~\ref{tab:app_human_eval_results}:


\begin{enumerate}
    \item From Table~\ref{tab:app_human_eval_results}, we can observe that the final rationale alignment is 72.0\% which means that final rationale obtained from COLLATE is reliable and aligns with human preferences.
    \item Rationales from the two rationale providers are diverse: It is observed that for 78.67\% cases, one rationale obtained was judged to be better than the other generated rationale. This means that employing two variants of same LLM is useful to obtain distinct and diverse rationales which are useful to improve quality of preference data for DPO.
    \item Likelihood based rationale selection aligns with human preferences: For 84.0\% cases, better generated rationale determined based on human preferences matches the better rationale based on likelihood-based utility score. This shows that our choice of using likelihood of final GT answer for selecting winner rationale aligns with human preferences and is suitable to obtain the preference data.
\end{enumerate}

\textbf{Inter-Annotator Agreement}: We also report the inter-annotator agreement by estimating the fleiss' kappa coefficient which is commonly used to measure agreement between three annotators. For the human study, the fleiss-kappa coefficient for inter-annotator agreement came out to be: 0.5829.

 
Following is mapping of fleiss-kappa coefficient value ranges with interpretation: 

0 – 0.2: Slight agreement \\
0.21 - 0.4: Fair agreement \\
0.41 - 0.6: Moderate agreement \\ 
0.61 - 0.8: Substantial agreement \\
0.81 - 1.0: Almost Perfect agreement \\ 

Based on the coefficient obtained for different metrics and the above scale, it can be seen that human labels have moderate to good agreement.

\section{Additional Related Work}
Differently, some works use mixture-of-experts where the instruction is routed to suitable LLM expert either at the query-level~\citep{lu-etal-2024-routing}, in latent space~\citep{jiang2024mixtralexperts} or at the output-layer~\citep{si-etal-2023-getting}.

\section{Justification for Splitting Data Randomly to Train Rationale Providers}
We intentionally divided the dataset into random splits for training and obtaining distinct rationale providers because of the consideration that if a rationale provider is trained on samples from a specific task, then it can become an expert in that task only and might not be able to generalize and generate useful rationale candidates for other tasks. However, obtaining diverse, good-quality and competing rationale candidates is critical to obtain a good rationales set for improving the rationale providers for a given task through preference tuning. Hence, by design, we divide the dataset randomly so as to prevent the rationale providers become very specific to a particular domain or task. However, we do agree that the effect of using domain-specialized rationale providers could be explored as future work.

\section{Comparison of \approachName\ with SPIN on 3 additional datasets}
To further demonstrate the robustness and generalizability of our proposed approach \approachName, we conducted evaluations on three additional, challenging datasets: \textbf{AIME} (a more recent and difficult benchmark for mathematical reasoning), \textbf{CROW} (common sense reasoning), and \textbf{HotpotQA} (long chain multi-hop reasoning). Evaluation on these datasets provide further empirical evidence for COLLATE's performance improvements over the strongest baseline, SPIN. These datasets complement the already diverse set of benchmarks used in the main paper, extending the coverage across: \textbf{1) Mathematical Reasoning:} AIME focuses on complex math word problems. \textbf{2) Commonsense Reasoning:} CROW requires applying background knowledge to infer correct outcomes. \textbf{3) Multi-hop QA:} HotpotQA evaluates long chain reasoning and contextual grounding.

The results are summarized in Table~\ref{tab:additional-results}, where we observe that COLLATE consistently outperforms SPIN across all three benchmarks, with an absolute improvement of \textbf{3.7\%} on AIME, \textbf{4\%} on CROW, and \textbf{1.7\%} on HotpotQA. These results strengthen our claim that COLLATE is a general-purpose reasoning enhancement approach that scales well across diverse and difficult reasoning tasks.

\begin{table}[t]
\centering
\caption{Performance comparison of COLLATE with SPIN on additional benchmarks.}
\label{tab:additional-results}
\resizebox{0.48\textwidth}{!}{%
\begin{tabular}{lccc}
\toprule
\textbf{Method} & \textbf{AIME} & \textbf{CROW} & \textbf{HotpotQA} \\
\midrule
SPIN & 16.02 & 68.29 & 62.39 \\
\approachName\ (ours) & \textbf{19.72} & \textbf{72.18} & \textbf{64.11} \\
\bottomrule
\end{tabular}}
\end{table}



\end{document}